\RequirePackage[2020-02-02]{latexrelease}
\documentclass{clv3_modified}

\usepackage{hyperref}
\usepackage{xcolor}
\definecolor{darkblue}{rgb}{0, 0, 0.5}
\hypersetup{colorlinks=true,citecolor=darkblue, linkcolor=darkblue, urlcolor=darkblue}

\usepackage{booktabs} 
\usepackage{multirow} 
\usepackage{makecell}

\makeatletter\newcommand{\tableofcontents}{\@starttoc{toc}}\makeatother
\begin{document}
\issue{1}{1}{2016}


\runningtitle{Gao, Hu, Yin, Ruan, Pu, and Wan \thepage}
\runningauthor{Survey Article}

\title{LLM-based NLG Evaluation: \\ Current Status and Challenges}

\author{Mingqi Gao$^{*}$}
\affil{Peking University \\
\texttt{gaomingqi@pku.edu.cn}
}

\author{Xinyu Hu$^{*}$}
\affil{Peking University \\
\texttt{huxinyu@pku.edu.cn}
}

\author{Xunjian Yin}
\affil{Peking University \\
\texttt{xjyin@pku.edu.cn}
}

\author{Jie Ruan}
\affil{Peking University \\
\texttt{ruanjie@stu.pku.edu.cn}
}

\author{Xiao Pu}
\affil{Peking University \\
\texttt{puxiao@stu.pku.edu.cn}
}

\author{Xiaojun Wan}
\affil{Peking University \\
\texttt{wanxiaojun@pku.edu.cn}
}

\maketitle
\def\thefootnote{*}\footnotetext{Equal contribution.}\def\thefootnote{\arabic{footnote}}

\begin{abstract}
Evaluating natural language generation (NLG) is a vital but challenging problem in natural language processing. Traditional evaluation metrics mainly capturing content (e.g. n-gram) overlap between system outputs and references are far from satisfactory, and large language models (LLMs) such as ChatGPT have demonstrated great potential in NLG evaluation in recent years. Various automatic evaluation methods based on LLMs have been proposed, including metrics derived from LLMs, prompting LLMs, fine-tuning LLMs, and human-LLM collaborative evaluation. In this survey, we first give a taxonomy of LLM-based NLG evaluation methods, and discuss their pros and cons, respectively. Lastly, we discuss several open problems in this area and point out future research directions.
\end{abstract}

\section*{Contents}
\tableofcontents

\bigskip

\section{Introduction}

The evaluation of natural language generation (NLG) is an important but challenging issue. The lack of a single standard answer and the presence of multiple quality criteria make evaluating NLG more challenging than other NLP tasks. For example, in news summarization, a good summary should capture the key information from the source document, remain faithful to the source document, and be expressed in logically coherent and fluent language, but there is not a single "correct" way to achieve this. The inherent difficulty of NLG evaluation means that human evaluation is always needed and regarded as the gold standard. However, due to the high cost and time-consuming nature of human evaluation, automatic evaluation metrics remain indispensable and play a crucial role in model development. Over the past two decades, many automatic evaluation metrics such as BLEU \citep{papineni2002bleu} and BARTScore \citep{yuan2021bartscore} have been developed, but none have been fully satisfactory. Some studies \citep{sai2021perturbation, he2023blind} have highlighted their deficiencies in robustness, such as insensitivities, biases, or even loopholes when evaluating challenging texts.

Recently, large language models (LLMs) have emerged and demonstrated unprecedented capacities in following instructions, understanding content, and generating text, which inspires researchers to utilize LLMs for NLG evaluation. Although this is a research direction that only emerged in 2023, the past year has seen an enormous amount of relevant work. While there have been surveys on automatic evaluation metrics or human evaluation practices in NLG evaluation \citep{DBLP:journals/corr/abs-2006-14799,DBLP:journals/corr/abs-2108-00308,DBLP:conf/naacl/ZhouBTDSO22,DBLP:journals/csur/SaiMK23,DBLP:journals/jair/GehrmannCS23,DBLP:conf/acl-clinicalnlp/0004RP23}, none of them addresses LLM-based evaluation approach, and a comprehensive survey of this area is urgently needed.

The survey will mainly focus on research on LLM-based approaches for NLG evaluation, which involves language models with over one billion parameters. Moreover, it mainly considers the typical scope of NLG tasks where both input and output are natural languages including machine translation, text summarization, story generation, dialogue response generation, data-to-text, text simplification, paraphrase generation, grammatical error correction, and creative writing. Broader areas like evaluation of LLMs are not included \citep{DBLP:journals/corr/abs-2308-07902,DBLP:journals/tist/ChangWWWYZCYWWYZCYYX24} because this work focuses on LLMs used for evaluation, rather than the evaluation of LLMs' capabilities. We search the literature on Google Scholar with an end date of June 2024 with keywords. Since this is a new direction that emerged in 2023, a considerable number of arXiv preprints are included in addition to papers published in *ACL venues or other related venues. About 100 pieces of work will be included. To maintain focus, this paper neither discusses datasets and benchmarks in NLG evaluation \citep{DBLP:journals/corr/abs-2102-01672,DBLP:journals/corr/abs-2406-05761} nor analyzes evaluation metrics statistically \citep{DBLP:conf/acl/NimahFMP23,DBLP:conf/emnlp/XiaoZLL23}.

\begin{figure}[h]
    \centering
    \includegraphics[width=0.85\linewidth]{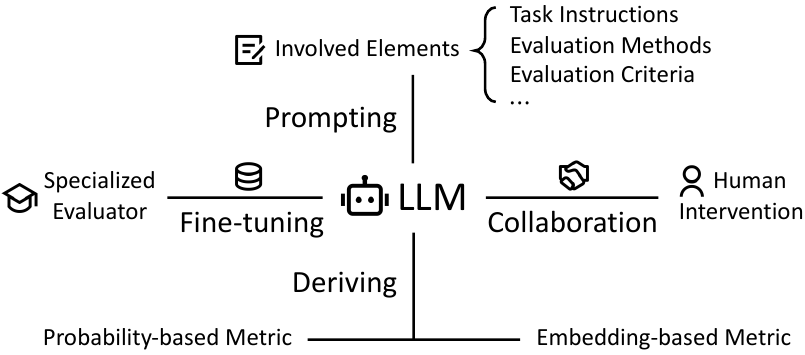}
    \caption{Schematic representation of our proposed four categories of LLM-based NLG evaluation.}
    \label{fig:enter-label}
\end{figure}

 As shown in \autoref{fig:enter-label}, we categorize related studies into four categories according to how LLMs are utilized for NLG evaluation:
\begin{itemize}
    \item \textbf{LLM-derived Metrics} (\S\autoref{sec:llm-derived}): developing or deriving evaluation metrics from embeddings or generation probabilities of LLMs.
    \item \textbf{Prompting LLMs} (\S\autoref{sec:prompting}): directly inquiring of existing LLMs via specific prompts and processes designed for evaluation. 
    \item \textbf{Fine-tuning LLMs}  (\S\autoref{sec:fine-tuning}): using labeled evaluation data to fine-tune existing LLMs and improving their NLG evaluation capabilities. 
    \item \textbf{Human-LLM Collaborative Evaluation} (\S\autoref{sec:collaboration}): leveraging distinctive strengths of both human evaluators and LLMs to achieve robust and nuanced evaluations through human-LLM collaboration.
\end{itemize}

\textbf{LLMs have driven NLG evaluation toward a more human-centered direction, and the four categories we propose reflect this evolution}: LLM-derived metrics are a continuation of traditional evaluation metrics and can only handle coarse-grained evaluation; prompting and fine-tuning methods enable users to express flexible evaluation requirements in natural language; collaborative evaluation takes it a step further, making it possible for humans and LLMs to leverage their strength respectively. We will review each type of evaluation method and discuss the pros and cons, respectively. Last but not least, we will provide our suggestions and conclusions, and discuss future directions in this area (\S\autoref{sec:future}). 

It is worth stating that since LLM-based evaluation has shown unprecedented generality across NLG tasks, we do not summarize the literature for each task separately. Nevertheless, we will draw a list documenting all the approaches in this survey, indicating which NLG tasks each approach has been experimented on.

\section{LLM-derived Metrics}
\label{sec:llm-derived}

LLM-derived metrics can be viewed as a continuation of early model-based NLG evaluation metrics such as BERTScore and BARTScore, replacing traditional pre-trained language models with stronger LLMs. Such works can be categorized into two main types: embedding-based metrics \citep{es2023ragas} and probability-based metrics. The latter can be further divided into two categories based on different ways of using probabilities: directly converting the probabilities into scores \citep{fu2023gptscore,varshney2023stitch} and leveraging the variation in probabilities under changed conditions \citep{jia2023zero,xie-etal-2023-deltascore}. \\

\subsection{Embedding-based Metrics} The embedding-based methods, like BERTScore, generally utilize representations of language models and thus compute the semantic similarity between the reference and the target text to evaluate, with different possible ways of implementation. 
However, unlike traditional embedding-based evaluation metrics that require references, many LLM-based embedding evaluation metrics do not. This is because their application scenarios and implementation methods differ from those of traditional metrics. For example, when \citet{es2023ragas} evaluate the answer relevance of Retrieval Augmented Generation, given the original question $q$ and the answer $Y$ to be evaluated, they first prompt the LLM to generate $n$ possible questions $q_i$ for $Y$. Then, the relevance of $Y$ is represented by the average similarity between $q_i$ and $q$, denoted as $\sum_{i=1}^n \text{sim}(q_i,q)$, where $\text{sim}(q_i,q)$ refers to the cosine similarity of the embeddings of $q_i$ and $q$. The embedding is generated by OpenAI \textit{text-embedding-ada-002}, which can efficiently convert text into a 1536-dimensional vector, capturing semantic information and ensuring that similar texts are positioned close to each other in the vector space. Furthermore, \citet{DBLP:conf/emnlp/Sheng0ZSFDZGWZ24} developed a more sophisticated method based on embeddings from the open-source decoder-only LLM, utilizing Principal Component Analysis to adapt it for both pointwise scoring and pairwise comparison.

\subsection{Probability-based Metrics} To better utilize the knowledge inherent in language models, probability-based methods like BARTScore formulate text generation evaluation as conditional probability comparison, positing that the better the quality of the target text, the higher the likelihood that models should be able to generate it. Recently, GPTScore \citep{fu2023gptscore} has established tailored evaluation templates for each aspect to effectively guide multiple LLMs for NLG evaluation, including GPT3 \citep{brown2020language}, OPT \citep{zhang2022opt}, and FLAN \citep{chung2022scaling}. The core idea of GPTScore is that a good generative language model is more likely to assign higher probabilities to high-quality text generated in response to a given instruction and context. Specifically, given a generative large language model $\theta$, context information $X$ (such as a source document), output text $Y = \{y_1, y_2, \dots, y_m\}$ containing $m$ tokens to be evaluated, and instruction $I$ that specifies the requirement for the LLMs to generate text that can flexibly correspond to different evaluation aspects (e.g., \textit{generating a factually consistent summary} for the aspect of consistency), GPTScore is defined as:
$$\text{GPTScore}(X, Y, I, \theta)=\sum_{i=1}^m \log P(y_i\vert y_{<i},X,I,\theta)$$

Similarly, \citet{murugadoss2024evaluating} score the task output $Y$ to be evaluated by its perplexity under the corresponding large language model $\theta$, given only the task context $X$. They believe this approach is unbiased by prompts, which transparently measures alignment with model training data. Furthermore, such methods have also been applied to the hallucination detection of the LLM-generated text \citep{varshney2023stitch} with three different attempts for calculating the probability score. 

On the other hand, some works leverage the variation in probabilities under changed conditions as the evaluation metric. FFLM \citep{jia2023zero} proposes to evaluate the faithfulness of the target text by calculating a combination of probability changes based on the intuition that the generation probability of a given text segment increases when more consistent information is provided, and vice versa. Similarly, DELTASCORE \citep{xie-etal-2023-deltascore} measures the quality of different story aspects according to the likelihood difference between pre- and post-perturbation states with LLMs including GPT-3.5 (text-davinci-003) that provide logits. They believe that the sensitivity to specific perturbations indicates the quality of related aspects, and their experiments demonstrate the effectiveness of their approach.

\subsection{Pros and Cons}

Traditional NLG evaluation approaches always fall short due to their surface-form similarity when the target text and reference convey the same meaning but use different expressions. In contrast, LLM-derived metrics offer a remedy for the limitation and demonstrate stronger correlations with human judgments benefiting from the evolving modeling techniques. However, the flaws within LLMs can lead to some issues, as introduced in the following:

\textbf{Robustness.} Some research has investigated the robustness of LLM-derived metrics and found that they lack robustness in different attack scenarios. Specifically, \citet{he-etal-2023-blind} develops a set of stress tests to assess the robustness of various model-based metrics on some common NLG tasks. They show a catalogue of the blind spots and potential errors identified that are not detected by different metrics.

\textbf{Efficiency.} Compared to traditional metrics, LLM-derived evaluation methods are more time-consuming and require more computational resources, especially when adopting LLMs with quite large parameter scales. To address this, \citet{eddine2022frugalscore} propose an approach to learning a lightweight version of LLM-derived metrics, and some fast LLM inference and serving tools like popular vLLM \citep{kwon2023efficient} have been launched. vLLM improves memory utilization during inference through the PagedAttention algorithm, as well as the optimized memory management and batching strategies, thereby increasing LLMs' generation throughput. However, closed-source LLMs often do not make their parameters, representations, or logits public and available, thus making it impossible to apply LLM-derived methods to them.

\textbf{Fairness.} \citet{sun2022bertscore} assess the social bias across various metrics for NLG evaluation on six sensitive attributes: race, gender, religion, physical appearance, age, and socioeconomic status. Their findings reveal that model-based metrics carry noticeably more social bias than traditional metrics. Relevant biases can be categorized into two types: intrinsic bias encoded within pre-trained language models and extrinsic bias injected during the computation of similarity. Therefore, current LLM-derived methods may have similar issues.

\section{Prompting LLMs}
\label{sec:prompting}

\begin{table*}
\renewcommand{\arraystretch}{1.35}
\setlength{\tabcolsep}{2.5pt}
\small
\centering
\begin{tabular}
{lccc}
\toprule
Related Work & Evaluation Method & NLG Task \\
\midrule
\citet{chiang-lee-2023-large} & Scoring & Story Generation \\
\citet{wang-etal-2023-chatgpt} & Scoring & \makecell[c]{Summarization, Data-to-text \\ \& Story Generation} \\
\citet{kocmi2023large} & Scoring & Translation \\
\citet{lin-chen-2023-llm} & Scoring & Dialogue \\
\citet{mendonca-etal-2023-simple}  & Scoring & Dialogue \\
\citet{naismith-etal-2023-automated}  & Scoring & Discourse Generation \\
\citet{liusie2023llm}  & \makecell[c]{Scoring \& \\Comparison} & \makecell[c]{Summarization, \\ Dialogue \& Data-to-text} \\
\citet{wang2023automated}  & Comparison & Personalized Text Generation \\
\citet{ji2023exploring}  & Ranking & Open-end Text Generation  \\
\citet{liu2023learning}  & \makecell[c]{Scoring, Ranking \\ \& Comparison} & Summarization \\
\citet{wang2023automatic}  & Boolean QA & Question Generation \\
\citet{manakul-etal-2023-selfcheckgpt}  & Boolean QA & Fact Verification \\
\citet{guan2023language}  & Boolean QA & Fact Verification \\
\citet{es2023ragas}  & Boolean QA & Retrieval Augmented Generation \\
\citet{kocmi-federmann-2023-gemba} & Error Analysis & Translation \\
\citet{lu2023error} & Error Analysis & Translation \\
\citet{chang2023booookscore} & Error Analysis & Summarization
\end{tabular}
\caption{Representative studies on prompting LLMs for NLG evaluation.}
\label{tab:overview}
\end{table*}

\begin{figure}[]
\centering
\includegraphics[width=0.95\textwidth]{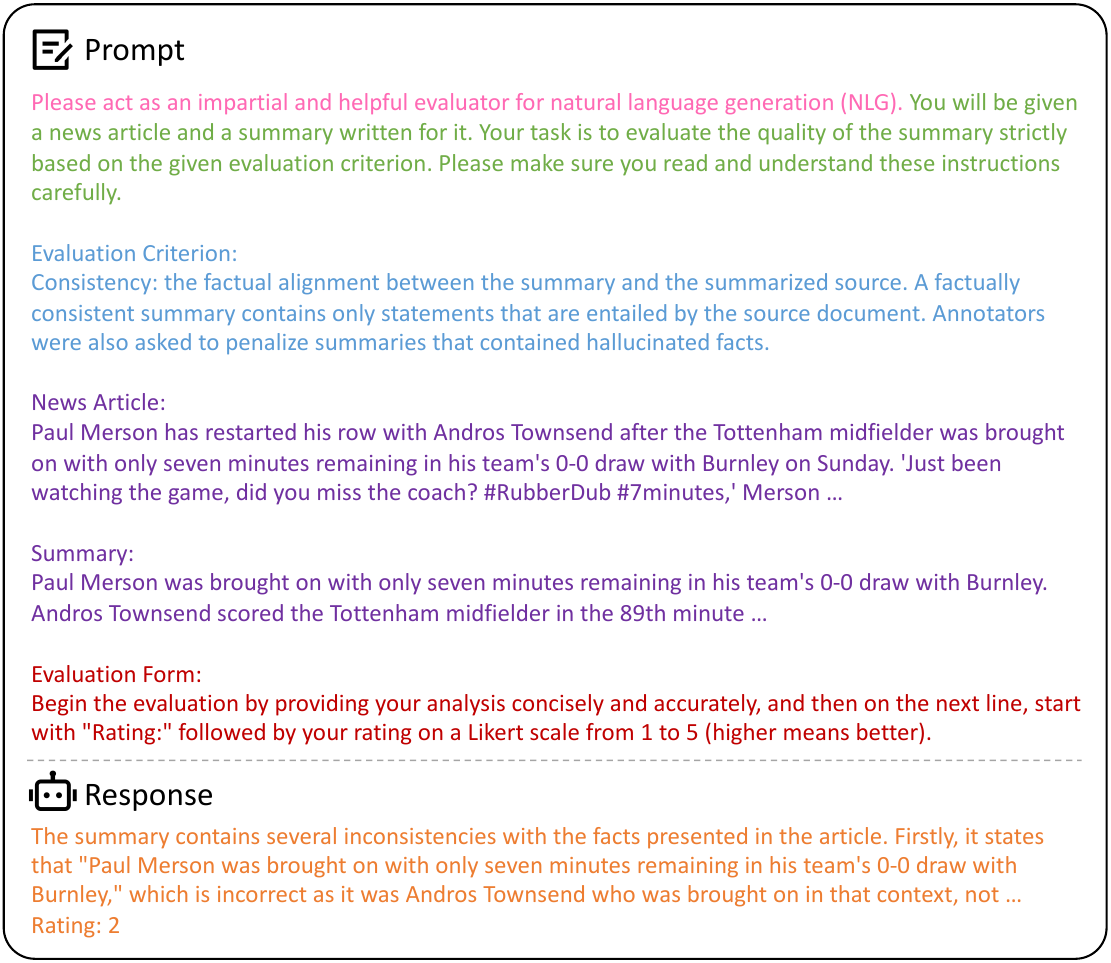}
\caption{An example of prompting LLMs to evaluate the aspect of consistency of the summary. There are \textcolor[RGB]{255,105,180}{role and interaction},  \textcolor[RGB]{112,173,71}{task instructions}, \textcolor[RGB]{91,155,213}{evaluation criteria}, \textcolor[RGB]{112,48,160}{input content}, and \textcolor[RGB]{192,0,0}{evaluation methods} in the prompt, as well as the \textcolor[RGB]{237,125,49}{evaluation results}, including the rating and explanation generated by LLMs.}
\label{fig:prompt}
\end{figure}

The remarkable generation abilities of LLMs have expanded the possibilities for NLG evaluation. For a long time, human evaluation has been viewed as the gold standard for NLG evaluation. Recently, some studies claim that LLMs are on par with crowdsourcing annotators in several tasks \citep{tornberg2023chatgpt,gilardi2023chatgpt,ostyakova-etal-2023-chatgpt,cegin-etal-2023-chatgpt}. This raises questions about whether LLMs could replace human evaluators. Studies in this area often involve feeding LLMs with detailed prompts that include both instructions and the text to be evaluated, with LLMs producing the evaluation outcomes. An example of prompting LLMs is shown in \autoref{fig:prompt}. From this example, we can see that such a prompt is quite similar to the guidelines given to human evaluators. The main differences between this prompting method for LLMs and LLM-derived metrics are twofold: (1) LLM-derived metrics generally do not involve highly human-like prompts that require the LLM to perform an evaluation. (2) The evaluation results from prompting LLMs are typically generated directly by the LLM, whereas LLM-derived metrics require further transformation from embeddings and probabilities. We will describe existing works according to the five elements that they mainly focus on:

\begin{itemize}
    \item \textbf{Evaluation Methods}: The way the evaluation results of LLM evaluators are obtained, such as scoring and comparison.
    \item \textbf{Task Instructions}: How LLM evaluators should read or manipulate different parts to complete the annotation.
    \item \textbf{Input Content}: The target text to be evaluated and other required content.  Other required content including source documents, references, and external knowledge is provided as needed.
    \item \textbf{Evaluation Criteria}: The general definition of how good or bad the text to be evaluated is in a particular aspect of quality, e.g. fluency, faithfulness.
    \item \textbf{Role and interaction}: The roles LLM evaluators play in the evaluation and the interactions between them.
\end{itemize}

\subsection{Evaluation Methods}

Diverse evaluation methods have been employed in prompting LLMs to obtain its preferences for the text to be evaluated: scoring, comparison, ranking, boolean QA, and error analysis (\autoref{tab:overview}).

\textbf{Scoring.} Scoring is the most commonly used evaluation method in human evaluation for NLG \citep{van2021human}, and it is naturally applied to LLM-based evaluation. \citet{chiang-lee-2023-large} have conducted relevant studies early, using a Likert scale from 1 to 5 to evaluate story generation and adversarial attacks with InstructGPT \citep{ouyang2022training} and ChatGPT \footnote{https://openai.com/blog/chatgpt/}, showing that the evaluation results of LLMs are consistent with expert human evaluators. \citet{kocmi2023large} discover GPT-3.5 and GPT-4 achieve the state-of-the-art accuracy of evaluating translation quality compared to human labels with a rating scale from 1 to 5 or 0 to 100, outperforming all the results from the metric shard task of WMT22 \citep{freitag-etal-2022-results}. Furthermore, \citet{wang-etal-2023-chatgpt} experiment on five datasets across summarization, story generation, and data-to-text, and ChatGPT with similar rating scales achieves the state-of-the-art or comparative correlations with human judgments in most settings, compared with prior metrics. Similar conclusions are also observed in open-domain dialogue response generation \citep{lin-chen-2023-llm}. Besides English, \citet{mendonca-etal-2023-simple} show that ChatGPT with simple rating prompts is a strong evaluator for multilingual dialogue evaluation, surpassing prior metrics based on encoders.

\textbf{Comparison.} Different from absolute scoring, comparison refers to choosing the better of the two.  \citet{luo2023chatgpt,gao2023humanlike} use ChatGPT to compare the factual consistency of two summaries. AuPEL \citep{wang2023automated} evaluate personalized text generation from three aspects in the form of comparison with the PaLM 2 family \citep{anil2023palm}. According to \citet{liusie2023llm}, pairwise comparison is better than scoring when medium-sized LLMs (e.g. FlanT5 \citep{chung2022scaling} and Llama2 \citep{touvron2023llama}) are adopted as evaluators.

\textbf{Ranking.} Ranking can be viewed as an extended form of comparison. In comparison, only two examples are involved at a time, whereas in ranking, the order of more than two examples needs to be decided at once. \citet{ji2023exploring} use ChatGPT to rank five model-generated responses across several use cases at once, indicating the ranking preferences of ChatGPT align with those of humans to some degree. Similarly, GPTRank is a method to rank summaries in a list-wise manner \citep{liu2023learning}. Moreover, \citet{liu2023benchmarking} compare different evaluation methods in LLM-based summarization including scoring, comparison, and ranking, showing that the optimal evaluation method for each backbone LLM may vary.

\textbf{Boolean QA.} Boolean QA requires LLMs to answer "Yes" or "No" to a question. It is adopted more in scenarios where human annotations are binary, such as grammaticality \citep{hu-etal-2023-decipherpref}, faithfulness of summaries and statements \citep{luo2023chatgpt,gao2023humanlike,es2023ragas,hu-etal-2023-decipherpref}, factuality of generated text \citep{fu2023large,guan2023language,manakul-etal-2023-selfcheckgpt}, and answerability of generated questions \citep{wang2023automatic}.

\textbf{Error Analysis.} Error Analysis refers to the evaluation of a text by looking for errors that occur in the text according to a set of predefined error categories. Multidimensional Quality Metrics (MQM) \citep{jain-etal-2023-multi} is an error analysis framework prevalent in machine translation evaluation. According to MQM, \citet{lu2023error,kocmi-federmann-2023-gemba} use ChatGPT or GPT-4 to automatically detect translation quality error spans. BOOOOKSCORE \citep{chang2023booookscore}, an LLM-based evaluation metric, assesses the coherence of book summaries by identifying eight types of errors.

\subsection{Task Instructions}
In human evaluation, task instruction usually comes in the form of a task description or evaluation steps. They can also exist at the same time. The task description states the annotation in a more general way, and the evaluation steps, which can be considered as Chain-of-Thought, explicitly describe what to do at each step. In the context of prompting LLMs for NLG evaluation, we discuss three broad categories of influences: various templates of prompts \citep{leiter2023eval4nlp,kim2023better,kotonya2023little,he2023socreval}, in-context examples \citep{jain-etal-2023-multi,kotonya2023little,hasanbeig2023allure}, and whether LLMs are required to provide analyses or explanations \citep{chiang-lee-2023-closer,naismith-etal-2023-automated}.

\textbf{Form and requirements.} Several studies from an Eval4NLP 2023 shared task \citep{leiter2023eval4nlp} have explored task instructions in various settings. For example, \citet{kim2023better} conduct experiments on different templates and lengths of task descriptions and evaluation steps, finding that providing clear and straightforward instructions, akin to those explained to humans, is more effective. 
\citet{kotonya2023little} generate task instructions with LLMs or improve existing task instructions with LLMs. Moreover, \citet{DBLP:conf/emnlp/LeiterE24} conduct a larger-scale prompt exploration for the evaluation of machine translation and summarization based on the Eval4NLP 2023 shared task. 
Somewhat differently, \citet{he2023socreval} evaluate generative reasoning using LLMs by asking them first to generate their own answers, and then conduct a quantitative analysis of the text to be evaluated. Additionally, explicit evaluation requirements and output formats are typically included in the instructions, and the evaluation results are extracted using regular expression matching. Early LLMs may sometimes provide unrecognizable evaluation results or refuse to conduct evaluation due to their limited instruction-following capabilities \citep{gao2023humanlike}. This issue can be mitigated through multiple sampling or setting random outputs, and it basically does not exist in the currently more advanced and powerful LLMs.

\textbf{Analysis and explanations.} LLMs are able to include analysis or explanation in their evaluations, which is a key point that distinguishes them from previous automatic evaluation metrics. Early explorations into prompting LLMs for NLG evaluation mostly do not examine the impact of whether LLMs are required to analyze and explain on evaluation results. However, \citet{chiang-lee-2023-closer} explore different types of evaluation instructions in summarization evaluation and dialogue evaluation, finding that explicitly asking large models to provide analysis or explanation achieve higher correlation with human judgments. Besides, the quality of the analysis and explanation generated by LLMs itself requires additional manual evaluation \citep{leiter2023eval4nlp}. \citet{naismith-etal-2023-automated} compare the explanations written by humans and generated by GPT-4 and conduct a simple corpus analysis on the generated explanations, finding that GPT-4 has strong potential to produce ratings that are comparable to human ratings on discourse coherence, accompanied by clear rationales.

\textbf{In-context examples.} Similarly to other fields, sometimes demonstrations are needed when prompting LLMs for NLG evaluation. Specifically, \citet{jain-etal-2023-multi} use only in-context examples as task instructions, relying on LLMs to evaluate the quality of summaries. In scenarios where task descriptions or evaluation steps are included, \citet{kotonya2023little} compare the performance of LLMs as evaluators in both zero-shot and one-shot settings, finding that one-shot prompting does not bring improvements. Moreover, \citet{hasanbeig2023allure} improve the performance of LLM evaluators by updating the in-context examples iteratively.

\subsection{Input Content}

The types of input content mainly depend on the evaluation criteria and are relatively fixed. For most task-specific evaluation criteria, such as the faithfulness of a summary \citep{luo2023chatgpt,gao2023humanlike}, the source document is needed in addition to the target text to be evaluated. For task-independent criteria, such as fluency \citep{hu-etal-2023-decipherpref,chiang-lee-2023-closer}, only the text to be evaluated needs to be provided, though many works also provide the source document \citep{wang-etal-2023-chatgpt,liusie2023llm}. Other types of input content can be provided as required by the specific task. \citet{kocmi2023large} use two different settings when evaluating machine translation: providing references and not providing references and find that GPT-4 without references can also outperform all existing reference-based metrics.
\citet{guan2023language} provide relevant facts and context when evaluating whether a text conforms to the facts. Exceptionally, \citet{shu2023fusioneval} add the output of other automatic evaluation metrics to the input of the LLM.

\subsection{Evaluation Criteria}

The evaluation targeting specific aspects is used in numerous studies of human evaluation for NLG, such as text summarization, story generation, dialogue, and text simplification. Evaluation criteria, i.e., the definitions of aspects are key in this context. Most evaluation criteria in LLM-based evaluation are directly derived from human evaluation. However, a few studies have attempted to let LLMs generate or improve evaluation criteria. \citet{liu2023calibrating} use a few human-rated examples as seeds to let LLMs draft some candidate evaluation criteria, and then further filter them based on the performance of LLMs using these criteria on a validation set, to obtain the final evaluation criteria. \citet{kim2023evallm} designed an LLM-based interactive evaluation system, which involves using LLMs to review the evaluation criteria provided by users, including eliminating ambiguities in criteria, merging criteria with overlapping meanings, and decomposing overly broad criteria. Additionally, \citet{ye2023flask} propose a hierarchical aspect classification system with 12 subcategories, demonstrating that under the proposed fine-grained aspect definitions, human evaluation and LLM-based evaluation are highly correlated. Besides, the chain-of-aspects approach improves LLMs' ability to evaluate on a specific aspect by having LLMs score on some related aspects before generating the final score \citep{gong2023coascore}.

\subsection{Role and Interaction}
We include in this section the evaluation strategies that either use the same LLMs in different ways or involve different LLMs \citep{bai2023benchmarking,li2023prd,cohen-etal-2023-lm}. The former can be further divided into chain-style \citep{yuan2023batcheval,fu2023large,hu-etal-2023-decipherpref} and network-style interactions \citep{chan2023chateval,zhang2023wider,saha2023branchsolvemerge,wu2023large}.

\textbf{Chain-style interaction.} Inspired by human evaluators, \citet{yuan2023batcheval} have LLMs score a batch of examples to be evaluated each time. Specifically, the evaluation process is divided into three stages: analysis, ranking, and scoring. Similar to QA-based evaluation metrics \citep{durmus-etal-2020-feqa}, \citet{fu2023large} assess the faithfulness of summaries in two stages: treating LLMs as question generators to generate a question from the summary; then having LLMs answer the question using the source document. Differently, when \citet{hu-etal-2023-decipherpref} use GPT-4 to evaluate the faithfulness of summaries, it first asks GPT-4 to extract event units from the summary, then verifies whether these event units meet the requirements, and finally judges whether the event units are faithful to the source document.

\textbf{Network-style interaction.} Unlike chain-style interactions, network-style interactions involve the dispersion and aggregation of information. In network-style interactions, LLMs on the same layer play similar roles. ChatEval \citep{chan2023chateval} is a framework for evaluating content through debates among multiple LLMs, with three communication strategies designed among the three types of LLMs: One-By-One, Simultaneous-Talk, and Simultaneous-Talk-with-Summarizer. \citet{zhang2023wider} find that under certain conditions, widening and deepening the network of LLMs can better align its evaluation with human judgments. \citet{saha2023branchsolvemerge} propose a branch-solve-merge strategy, assigning LLMs the roles of decomposing problems, solving them, and aggregating answers, thereby improving the accuracy and reliability of evaluations. \citet{wu2023large} assume that different people such as politicians and the general public have different concerns about the quality of news summaries, use LLMs to play different roles in evaluation accordingly, and aggregate the results finally.

\textbf{Different LLMs.} Different from having the same LLM play different roles, some research has used different LLMs (such as GPT-4 and Claude) in their studies. The use of a single LLM as evaluator may introduce bias, resulting in unfair evaluation results. In light of this, \citet{bai2023benchmarking} design a decentralized Peer-examination method, using different LLMs as evaluators and then aggregating the results. Further, \citet{li2023prd} let different LLMs serve as evaluators in pairwise comparisons and then have them go through a round of discussion to reach the final result. Additionally, \citet{cohen-etal-2023-lm} evaluate the factuality of texts through the interaction of two LLMs, where the LLM that generated the text acts as the examinee and the other LLM as the examiner.

\subsection{Pros and Cons}

The benefits of prompting LLMs for NLG evaluation are exciting. First, for the first time, people can express evaluation criteria and evaluation methods in natural language within the prompts given to LLMs, providing great flexibility. Where previously people needed to design specific evaluation metrics for different NLG tasks or even different aspects of a single task, now they only need to modify the prompts for LLMs. Secondly, surprisingly, LLMs have the ability to generate explanations while assessing texts, making this approach somewhat interpretable. Furthermore, in many NLG tasks, prompting LLMs for evaluation has achieved state-of-the-art correlations with human judgments.

However, as many studies have pointed out, this type of approach still has many limitations. \citet{wang2023large} note that when using ChatGPT and GPT-4 for pairwise comparisons, the order of the two texts can affect the evaluation results, which is known as position bias. To alleviate this issue, \citet{li2023split} propose a strategy of splitting, aligning, and then merging the two texts to be evaluated into the prompt. Also, LLM evaluators tend to favor longer, more verbose responses \citep{zheng2023judging} and responses generated by themselves \citep{liu-etal-2023-g}.  \citet{wu2023style} show that compared to answers that are too short or grammatically incorrect, answers with factual errors are considered better by LLMs. \citet{liu2023evaluate} demonstrate through adversarial meta-evaluation that LLMs without references are not suitable for evaluating dialogue responses in closed-ended scenarios: they tend to score highly on responses that conflict with the facts in the dialogue history. \citet{zhang2024comprehensive} also present the robustness issues of LLMs in dialogue evaluation through adversarial perturbations. \citet{shen-etal-2023-large} indicate that LLM evaluators have a lower correlation with human assessments when scoring high-quality summaries. In addition, \citet{hada2023large} state that LLM-based evaluators have a bias towards high scores, especially in non-Latin languages like Chinese and Japanese. \citet{DBLP:journals/corr/abs-2406-18403} find that the performance of LLM-based evaluators exhibits significant variance depending on the dataset, evaluation criteria, and whether the evaluated texts are human-generated. Beyond these shortcomings of performance, both ChatGPT and GPT-4 are proprietary models, and their opacity could lead to irreproducible evaluation results.

\section{Fine-tuning LLMs}
\label{sec:fine-tuning}

As mentioned above, despite the exciting performance of prompting LLMs like ChatGPT and GPT-4 for NLG evaluation, several shortcomings in practice are inevitable, such as high costs, possibly irreproducible results, and potential biases in LLMs. In response, recent research has shifted towards fine-tuning smaller, open-source LLMs specifically for evaluation purposes, aiming to achieve performance close to GPT-4 in NLG evaluation. Representative works of this type include PandaLM \citep{wang2023pandalm}, Prometheus \citep{kim2023prometheus}, Prometheus 2 \citep{kim-etal-2024-prometheus}, Shepherd \citep{wang2023shepherd}, TIGERScore \citep{jiang2023tigerscore}, INSTRUCTSCORE \citep{xu-etal-2023-instructscore}, Auto-J \citep{li2023generative}, CritiqueLLM \citep{ke2023critiquellm}, JudgeLM \citep{zhu2023judgelm}, Themis \citep{hu-etal-2024-themis}, CompassJudger-1 \citep{cao2024compassjudger} and Self-Taught \citep{wang2024self}. Their main ideas are similar, involving the elaborate construction of high-quality evaluation data, followed by fine-tuning open-source foundation LLMs with specific methods. Nevertheless, there are certain discrepancies in the designs across different works, such as the usage of references and evaluation criteria. We have summarized the key different components of these methods in \autoref{tab:comparison1} and \autoref{tab:comparison2} for comparison, which will be elaborated in the following sections.

\begin{table*}
\small
\setlength{\tabcolsep}{1.5pt}
\renewcommand{\arraystretch}{1.25}
\centering
\begin{tabular}
{c@{\hspace{10pt}}c@{\hspace{10pt}}c@{\hspace{18pt}}c@{\hspace{18pt}}c}
\toprule
\multirow{2.5}{*}{Method} & \multicolumn{3}{c}{Data Construction} & \multirow{2.5}{*}{Foundation LLM} \\
\cmidrule(lr){2-4} & Instruction Source & Annotator & Scale \\
\midrule
PandaLM & Alpaca 52K & GPT-3.5 & 300K & \makecell[c]{LLaMA 7B} \\[3pt]

Prometheus & GPT-4 Construction & GPT-4 & 100K & \makecell[c]{LLaMA-2-Chat \\7B \& 13B} \\[6pt]

Prometheus 2 & FEEDBACK COLLECTION & GPT-4 & 200K & \makecell[c]{Mistral-7B \\ Mixtral-8x7B} \\[6pt]

Shepherd & \makecell[c]{Community Critique Data \\ \& 9 NLP Tasks Data} & Human & 1317 & \makecell[c]{LLaMA 7B} \\[6pt]

TIGERScore & \makecell[c]{23 Distinctive Text \\ Generation Datasets} & GPT-4 & 48K & \makecell[c]{LLaMA-2 \\7B \& 13B} \\[6pt]

INSTRUCTSCORE & GPT-4 Construction & GPT-4 & 40K & \makecell[c]{LLaMA 7B} \\[6pt]

AUTO-J & \makecell[c]{Real-world User Queries \\ from Preference Datasets} & GPT-4 & 4396 & \makecell[c]{LLaMA-2-Chat\\13B} \\[6pt]

CritiqueLLM & \makecell[c]{AlignBench \& \\ChatGPT Augmentation} & GPT-4 & 9332 & \makecell[c]{ChatGLM-2 \\6B, 12B \& 66B} \\[6pt]

JudgeLM & \makecell[c]{GPT4All-LAION, ShareGPT \\ Alpaca-GPT4 \& Dolly-15K} & GPT-4 & 100K & \makecell[c]{Vicuna \\7B, 13B \& 33B} \\[6pt]

Themis & \makecell[c]{NLG-Eval with 58 NLG\\ Evaluation Datasets} & \makecell[c]{Human \\ \& GPT-4} & 67K & \makecell[c]{Llama-3-8B} \\[6pt]

Self-Taught & \makecell[c]{Screened WildChat} & \makecell[c]{Llama-3-70B} & 20K & \makecell[c]{Llama-3-70B} \\[6pt]

CompassJudger-1 & \makecell[c]{Sampling from \\ existing datasets} & \makecell[c]{Mixture} & 900K & \makecell[c]{Qwen-2.5 1.5B, \\ 7B, 14B \& 32B} \\[6pt]

\end{tabular}
\caption{Comparison of the different key components among the representative methods of fine-tuning LLMs (Part 1).}
\label{tab:comparison1}
\renewcommand{\arraystretch}{1}
\end{table*}

\begin{table*}
\small
\renewcommand{\arraystretch}{1.2}
\centering
\begin{tabular}
{c@{\hspace{10pt}}c@{\hspace{10pt}}c@{\hspace{10pt}}c@{\hspace{10pt}}c}
\toprule
\multirow{2.5}{*}{Method} & \multicolumn{3}{c}{Evaluation Method} & \multirow{2.5}{*}{\makecell[c]{Reference \\Required}} \\
\cmidrule(lr){2-4} & Result Mode & Details & Specific Criteria \\
\midrule
PandaLM & Comparison & \makecell[c]{Reason \& Reference} & Unified & No \\[3pt]

Prometheus & Scoring & Reason & Explicit & Yes \\[6pt]

Prometheus 2 & \makecell[c]{Scoring \& \\ Comparison} & Reason & Explicit & Yes \\[6pt]

Shepherd & \makecell[c]{Overall Judgement} & \makecell[c]{Error Identifying \\ \& Refinement} & Unified & No \\[6pt]

TIGERScore & MQM & Error Analysis & Implicit & No \\[6pt]

INSTRUCTSCORE & MQM & Error Analysis & Implicit & Yes \\[6pt]

AUTO-J & \makecell[c]{Scoring \& \\ Comparison} & Reason & Implicit & No \\[6pt]

CritiqueLLM & Scoring & Reason & Unified & Flexible \\[6pt]

JudgeLM & \makecell[c]{Scoring \& \\ Comparison} & Reason & Unified & Flexible \\[6pt]

Themis & Scoring & Reason & Explicit & No \\[6pt]

Self-Taught & Comparison & Reason & Unified & No \\[6pt]

CompassJudger-1 & \makecell[c]{Scoring \& \\ Comparison} & Reason & Explicit & No \\[6pt]

\end{tabular}
\caption{Comparison of the different key components among the representative methods of fine-tuning LLMs (Part 2).}
\label{tab:comparison2}
\end{table*}

\subsection{Data Construction}

Diverse data with high-quality annotations is crucial for the fine-tuning of evaluation models, which mainly involves task scenarios, inputs, target texts to evaluate, and evaluation results. Early NLG evaluation research primarily focused on conventional NLG tasks, such as summarization and dialogue generation. Thus, the task scenarios, inputs, and target texts refer to the corresponding NLP task, source inputs of the task, and outputs generated by specialized systems based on task requirements, respectively. And mainstream datasets for these tasks predominantly employ human annotators to provide evaluation results, which are often considered reliable.

With the recent rise of LLMs, the spectrum of NLG tasks has been broadened to scenarios of instruction and response that are more aligned with human needs. Traditional tasks like summarization with corresponding source inputs can be viewed as kinds of instructions and requirements. Meanwhile, responses generated by various general LLMs generally serve as the target texts now and require more flexible evaluation so that the performance of different LLMs can be compared, promoting further developments. Therefore, to keep pace with the current advancement of modeling techniques, most evaluation methods have adopted the similar instruction-response scenario.

The primary differences in these works actually lie in the construction of instructions, with the purpose of improving either diversity or reliability for the better generalization ability of the fine-tuned model. PandaLM and JudgeLM entirely sample from common instruction datasets, such as Alpaca 52K, while CritiqueLLM adopts small-scale sampling followed by ChatGPT augmentation. In contrast, Prometheus and INSTRUCTSCORE rely on GPT-4 to generate all the instructions based on seed data, whereas Auto-J and Shepherd use real-world data. Moreover, since large-scale human annotation is impractical, most works utilize GPT-4 as the powerful annotator, except for PandaLM and Shepherd, which use GPT-3.5 and human annotation on small-scale community data, respectively. Specifically, Themis focuses on NLG tasks and combines existing human evaluations with additional evaluations from GPT-4, selecting more consistent training data. Self-Taught uses the evaluation results from the model to fine-tune itself (Llama-3-70B), considering it already possesses strong capabilities. During the construction, these studies basically all design detailed prompts or guidance and apply heuristic filtering strategies and post-processing methods to mitigate noise. Overall, despite the possible higher quality of human annotation, the corresponding drawback is the difficulty in constructing large-scale datasets, which in turn may hinder adequate model training, while using LLMs for construction is the opposite situation.

\subsection{Evaluation Method}
As with prompting LLMs, the evaluation methods adopted in these works are highly diversified, involving different evaluation criteria, result modes, and usages of the reference. Given that current instruction-response scenarios encompass different types of tasks, it is unsuitable to specify unified evaluation criteria as in traditional NLG tasks. However, some works still do it this way, while some other methods let LLM annotators adaptively and implicitly reflect the required criteria in their evaluations, like PandaLM, TIGERScore, and AUTO-J. In particular, AUTO-J has meticulously crafted 332 evaluation criteria, matched to different tasks. Furthermore, Prometheus and Themis explicitly incorporate evaluation criteria into the evaluation instructions, and CompassJudger can work either with or without evaluation criteria, enabling flexible evaluation based on various customized criteria.

More details about the evaluation methods are shown in \autoref{tab:comparison2}. All the works require models to provide detailed information, such as reasons for their evaluation results. And the MQM mode can achieve more informative error analysis, offering stronger interpretability. Moreover, some works do not necessarily require references and then have greater value in practice. And a more optimal method is to concurrently support both reference-based and reference-free evaluations as JudgeLM and CritiqueLLM.

\subsection{Fine-tuning Implementation}
The fine-tuning process is implemented by different studies on their selected open-source foundation LLMs, like LLaMA, and respective constructed data, with some targeted settings. Specifically, Prometheus maintains balanced data distributions during fine-tuning, including the length and label. JudgeLM eliminates potential biases by randomly swapping sample pairs to be compared and randomly removing references. INSTRUCTSCORE utilizes GPT-4 to provide error annotations for the intermediate outputs of the fine-tuned model for further supervised reinforcement. And based on some preliminary experiments and manual analysis, TIGERScore determines appropriate ratios of different types of data during fine-tuning, which are claimed to be crucial by them. Moreover, CritiqueLLM implements separately, with and without references, and explores the effects of data and model scale. Themis employs additional rating-guided preference optimization after the fine-tuning process. Specifically, Self-Taught utilizes the evaluation results of the fine-tuned model itself for self-iterative optimization, leading to surprising improvements. Compared to the vanilla fine-tuning setting, these methods have improved the efficiency of model training and the robustness of evaluations.

\subsection{Pros and Cons}

The shortcomings of prompting LLMs for NLG evaluation can be significantly alleviated by the customized construction of training data and specifically fine-tuned LLMs. For instance, most models in \autoref{tab:comparison1} have less than 14B parameters, facilitating low-cost inference in practice and good reproducibility, with performance comparable to GPT4. And specific measures can be adopted to prevent certain biases found in GPT4 during different stages, such as randomly changing the order of training pairs for position bias. Furthermore, this type of approach allows for continuous iteration and improvement of the model to address potential deficiencies or emerging issues discovered in future applications.

However, some inherent biases associated with GPT4 may still persist, like self-biases, as the data construction of most methods employs GPT4 for critical evaluation annotation. On the other hand, many studies have chosen open-source foundation LLMs spanning three generations of the Llama series. With the recent rapid updates and improvements of open-source LLMs, it is intuitive that employing a more powerful foundation LLM should lead to better evaluation performance of the fine-tuned model. However, this means repetitive fine-tuning processes and computational expenses from scratch since directly migrating existing fine-tuned models to the new foundation LLM is challenging.

Additionally, although many existing methods aspire to more flexible and comprehensive evaluation through fine-tuning, demanding excessive evaluation settings may ultimately lead to poor performance or failure in model training, as AUTO-J and CritiqueLLM were found to have difficulties with criteria and references, respectively. However, there are some disagreements here since Prometheus, JudgeLM, and CompassJudger show different results, indicating such a seemingly straightforward fine-tuning process is actually quite complex. Moreover, considering the different evaluation settings in existing works, conducting a horizontal comparison among them is challenging. These issues require further exploration in future research.

\section{Human-LLM Collaborative Evaluation}
\label{sec:collaboration}

Human evaluation remains the gold standard for NLG due to its ability to capture nuanced aspects of quality.  However, it is expensive, time-consuming, and prone to subjective biases \citep{van2021human,deriu2021survey,li2023collaborative}.  While LLMs offer a promising avenue for automated evaluation, their reliability and correlation with human judgment are still areas of active development \citep{li2023split,liu2023evaluate}.  Human-LLM collaborative evaluation seeks to leverage the strengths of both: the nuanced judgment of humans and the scalability and efficiency of LLMs.
This chapter explores emerging paradigms in this collaborative space, focusing on how humans and LLMs can work together to improve the accuracy, efficiency, and trustworthiness of NLG evaluation. This includes collaborative approaches like: traditional evaluation tasks such as scoring and explaining \citep{zhang2021human,li2023collaborative}; general evaluation tasks such as testing and debugging \citep{ribeiro2022adaptive}; auditing NLG models to ensure fairness \citep{rastogi2023supporting}; aligning LLM-assisted evaluation of LLM outputs with human preferences \citep{10.1145/3654777.3676450}; addressing the intricate challenge of scalable oversight \citep{amodei2016concrete,saunders2022self}.


\subsection{Human-Guided LLM Evaluation}
Some works \citep{zhang2021human,li2023collaborative} focus on approaches where LLMs perform the primary evaluation task, but with significant guidance and oversight from humans. This guidance can take several forms, from designing detailed evaluation criteria to refining LLM outputs.

One common method is called checklist-based evaluation. A key challenge in open-ended NLG tasks is the lack of consistent evaluation criteria.  \citet{li2023collaborative} address this with COEVAL, a collaborative pipeline where humans design a task-specific checklist.  LLMs then use this checklist to generate initial evaluations and explanations, drawing on developments in explainable NLP \citep{yin2022interpreting,jung2022maieutic,ribeiro2022adaptive,ye2022complementary}. Humans then scrutinize these LLM-generated evaluations, refining scores and explanations.  This approach leverages the LLM's ability to process large amounts of text while retaining human oversight to ensure accuracy and reduce outliers.  Notably, human review still leads to revisions in approximately 20\% of LLM scores, highlighting the importance of human judgment.
Furthermore, InteractEval \citep{chu2025thinkworkbettercombining} combines human and LLM-generated attributes using Think Aloud methods to create questions and produce final prediction scores.
Think Aloud methods mean that human experts verbalize their thoughts and LLMs articulate their knowledge to generate text attribute insights using sample texts and evaluation rubrics, which highlights the necessity of effectively combining humans and LLMs in an automated checklist-based text evaluation.

Collaborative assignment is also useful for human-guided LLM evaluation.
\citet{zhang2021human} propose HMCEval, a framework that frames dialogue evaluation as a sample assignment problem.  This approach aims to optimize the allocation of evaluation tasks between humans and machines to maximize accuracy while minimizing human effort.  HMCEval achieves high accuracy (99\%) with significantly reduced human involvement (half the effort).
Besides, EvalAssist \citep{ashktorab2024aligninghumanllmjudgments} can help practitioners refine evaluation criteria using both direct and pairwise assessment strategies. \citet{ashktorab2024aligninghumanllmjudgments} also examine how users refine their criteria and identify key differences between the two evaluation approaches examined how users refine their criteria and identified key differences between the two evaluation approaches.

\subsection{LLM-Assisted Human Evaluation}
Some works \citep{ribeiro2022adaptive,rastogi2023supporting,POZDNIAKOV2024100289} explore scenarios where humans remain the primary evaluators, but LLMs provide assistance to improve efficiency, identify flaws, or audit for biases.

\citet{ribeiro2022adaptive} introduce AdaTest, a system where LLMs generate unit tests to identify bugs in a target NLG model.  Human feedback guides the LLM, significantly increasing the effectiveness of bug detection (5-10x improvement). This demonstrates the power of LLMs in generating diverse test cases, guided by human intuition.
In the task of evaluating machine translation systems, \citet{zouhar2025aiassistedhumanevaluationmachine} assist annotators by pre-filling error annotations with recall-oriented automatic quality estimation, which achieves the effect of reducing the time per span annotation by half while maintaining the same annotation quality level and further cutting the annotation budget by almost 25\%.

Addressing biases and irresponsible behavior in LLMs is crucial \citep{blodgett2020language,jones2022capturing}.  AdaTest++ \citep{rastogi2023supporting}, drawing on human-AI collaboration research, facilitates collaborative auditing.  Humans leverage their strengths in schematization and hypothesis testing, while LLMs assist in identifying a wide range of failure modes.  This collaborative approach uncovered both previously known and under-reported issues.

Evaluating LLMs on complex tasks can be challenging even for humans \citep{chen2021evaluating,nakano2021webgpt,li2022competition,menick2022teaching}.  The concept of scalable oversight \citep{amodei2016concrete} suggests using AI to assist in evaluation.  \citet{saunders2022self} explore using LLM-generated critiques to help humans identify flaws in model outputs, demonstrating that this form of assistance improves human performance.
What's more, \citet{POZDNIAKOV2024100289} focus on designing conversational user interfaces, which helps educators to use LLMs to evaluate assignments of students.

\subsection{Pros and Cons}
Human-LLM collaborative evaluation offers a compelling balance between the accuracy of human judgment and the efficiency of automated methods. Key advantages include:
(1) \textbf{Efficiency and Cost-Effectiveness}:  LLMs can significantly reduce the time and resources required for evaluation.
(2) \textbf{Complementary Strengths}:  Humans excel at nuanced judgment and critical thinking, while LLMs excel at processing large amounts of data and generating diverse outputs.
(3) \textbf{Improved Accuracy}:  Combining human and LLM strengths can lead to more accurate and reliable evaluations than either approach alone.

However, challenges remain:
(1) \textbf{Prompt Sensitivity}: LLM evaluation results can be sensitive to the phrasing of prompts, requiring careful prompt engineering \citep{li2023collaborative,rastogi2023supporting}.
(2) \textbf{Confidence Calibration}:  LLMs' ability to accurately assess their own confidence is still limited, making it difficult to know when to trust their judgments.
(3) \textbf{Need for Human Oversight}:  While reduced, human supervision is still necessary, limiting the potential for full automation.
(4) \textbf{Explainability}: Ensuring the collaborative process is transparent and understandable can be challenging.

\begin{table*}
\centering
\small
\setlength{\tabcolsep}{3.2pt}
\renewcommand{\arraystretch}{1.25}
\begin{tabular}{lcccccc}
\toprule
Method & Parameter & Coherence & Consistency & Fluency & Relevance & Overall \\

\midrule
\multicolumn{7}{l}{\emph{Traditional Metrics}} \\

BERTScore & 355M & 0.285  & 0.151 & 0.186 & 0.302 & 0.231 \\
BARTScore & 400M & 0.474 & 0.266 & 0.258 & 0.318 & 0.329 \\

\midrule

\multicolumn{7}{l}{\emph{LLM-derived Metrics}} \\

GPTScore (FT5) & 11B & 0.456 & 0.438 & 0.424 & 0.343 & 0.415 \\
GPTScore (OPT) & 66B & 0.359 & 0.453 & 0.380 & 0.337 & 0.382\\
GPTScore (GPT-3) & 175B & 0.434 & 0.449 & 0.403 & 0.381 & 0.417\\
GPTScore (Phi-4) & 14B & 0.319 & 0.436 & 0.386 & 0.154 & 0.324 \\
GPTScore (Llama-3.1) & 70B & 0.415 & 0.478 & 0.437 & 0.288 & 0.405 \\
GPTScore (Qwen-2.5) & 72B & 0.447 & 0.486 & 0.437 & 0.376 & 0.436 \\

\midrule
\multicolumn{7}{l}{\emph{Prompting LLMs}} \\

G-Eval (GPT-3.5) & - & 0.440 & 0.386 & 0.424 & 0.385 & 0.409 \\
G-Eval (GPT-4) & - & 0.582 & 0.507 & 0.455 & 0.548 & 0.523 \\
Phi-4 & 14B & 0.479 & 0.454 & 0.421 & 0.452 & 0.451 \\
Llama-3.1 & 70B & 0.510 & 0.387 & 0.317 & 0.494 & 0.427 \\
Qwen-2.5 & 72B & 0.515 & 0.509 & 0.435 & 0.528 & 0.497 \\

\midrule
\multicolumn{7}{l}{\emph{Fine-tuning LLMs}} \\
INSTRUCTSCORE & 7B & 0.328 & 0.232 & 0.260 & 0.211 & 0.258  \\
Prometheus 2 & 7B & 0.403 & 0.318 & 0.269 & 0.356 & 0.336 \\ 
Themis & 8B & 0.566 & 0.600 & 0.571 & 0.474 & 0.553 \\
TIGERScore & 13B & 0.381 & 0.427  & 0.363  & 0.366  & 0.384 \\
CompassJudger-1 & 32B & 0.494 &	0.424 &	0.318 &	0.410 &	0.411 \\ 

\midrule
\multicolumn{7}{l}{\emph{Human-LLM Collaborative Evaluation}} \\

InteractEval (GPT-3.5 1st) & - & 0.583 & 0.630 & 0.734 & 0.614 & 0.640 \\
InteractEval (GPT-3.5 2nd) & - & 0.590 & 0.614 & 0.726 & 0.623 & 0.638 \\
InteractEval (GPT-4 1st) & - & 0.649 & 0.799 & 0.783 & 0.626 & 0.714 \\
InteractEval (GPT-4 2nd) & - & 0.660 & 0.781 & 0.816 & 0.642 & 0.725 \\

\end{tabular}
\caption{Performance of different types of LLM-based NLG evaluation approaches on SummEval, where some results are from \citet{fu2023gptscore}, \citet{hu-etal-2024-themis} and \citet{chu2025thinkworkbettercombining}.}
\label{tab:Summarization}
\end{table*}

\section{Conclusions and Future Trends}
\label{sec:future}

\subsection{Comparison with traditional evaluation metrics.}
Traditional evaluation metrics are criticized for their poor correlation with human judgments \citep{DBLP:conf/cicling/StentMS05}, uninterpretable evaluation results \citep{DBLP:conf/lrec/ZhangVW04}, and inability to adapt to specific evaluation criteria \citep{DBLP:conf/emnlp/WisemanSR17}, which are being greatly mitigated by LLM-based evaluation. However, the higher cost, the requirements for computing resources, and the issues of reproducibility may be the downside. 

\subsection{Comparison between different types of LLM-based NLG evaluation.}

We compare different types of LLM-based evaluation according to flexibility and reproducibility due to the difficulty of comparing the effectiveness of different types of methods in various scenarios.

\paragraph{\textbf{Flexibility}} Human-LLM Collaborative Evaluation > Prompting LLMs > Fine-tuning LLMs > LLM-derived Metrics. Human-LLM Collaborative Evaluation involves human annotators, which provides the highest flexibility. LLM-derived Metrics are typically designed to evaluate specific aspects, such as text similarity, and do not fully allow evaluation criteria to be expressed in natural language, making them the least flexible. When comparing Prompting LLMs and Fine-tuning LLMs, the former, which uses proprietary models, generally performs better at following instructions compared to smaller open-source models.

\paragraph{\textbf{Reproducibility}} LLM-derived Metrics $\approx$ Prompting LLMs > Fine-tuning LLMs > Human-LLM Collaborative Evaluation. Human-LLM Collaborative Evaluation requires human annotators, and the recruitment and training of these annotators pose greater challenges to reproducibility. LLM-derived Metrics and Prompting LLMs do not modify the existing models, and therefore have better reproducibility than Fine-tuning LLMs. However, they may still become non-reproducible if proprietary models are deprecated.

\paragraph{\textbf{Performance}} Human-LLM Collaborative Evaluation > Fine-tuning LLMs $\approx$ Prompting LLMs > LLM-derived Metrics. We compare the performance of different LLM-based evaluation approaches on the most commonly used NLG evaluation benchmark on summarization, SummEval \citep{fabbri2021summeval}, as shown in \autoref{tab:Summarization}. When using the same LLMs, LLM-derived metrics perform worse than directly prompting LLMs for evaluation, and the latter is more convenient. Moreover, among methods of fine-tuning LLMs, only models focused on NLG evaluation scenarios, such as Themis, outperform prompting-based methods, including that using GPT-4. Other studies either use relatively outdated foundation LLMs or lack training on specific evaluation aspects like those in SummEval, leading to relatively weaker performance. Furthermore, Human-LLM Collaborative Evaluation enhances the LLM evaluation by incorporating checklists elaborated with human expert insights and LLM knowledge, resulting in the strongest performance.

\paragraph{\textbf{Cost}} LLM-derived Metrics $\approx$ Prompting open-source LLMs < Fine-tuning LLMs $\approx$ Prompting proprietary LLMs < Human-LLM Collaborative Evaluation. When using the same open-source LLM, the inference costs of LLM-derived metrics, prompting LLM, and fine-tuning LLM methods are the same, while fine-tuning LLM incurs additional training costs. When prompting proprietary LLMs, the cost is high and mainly concentrated in API calls during evaluation, making it difficult to directly compare with the training cost required for fine-tuning LLM. Moreover, human-LLM collaborative evaluation requires the involvement of human experts for each task, making it the most expensive approach.

\subsection{Future Directions}

\textbf{Unified benchmarks for LLM-based NLG evaluation approaches.} As mentioned above, each of the studies that fine-tuned LLMs to construct specialized evaluation models uses different settings and data during testing, making them incomparable. In the research on prompting LLMs for NLG evaluation, there are some publicly available human judgments on the same NLG task, such as SummEval for summarization. However, the existing human judgments have many problems. Firstly, most of the existing data only involve one type of NLG task and a single human evaluation method (e.g., scoring), making it difficult to evaluate LLMs' performance on different tasks, as well as using different evaluation methods on the same task. Secondly, many of the texts in these human judgments are generated by outdated models (such as Pointer Network) and do not include texts generated by more advanced LLMs. Lastly, many human evaluation datasets are too small in scale. There is an urgent need for large-scale, high-quality human evaluation data covering various NLG tasks and evaluation methods as a benchmark.

\textbf{NLG evaluation for low-resource languages and new task scenarios.} Almost all existing research focuses on English data. However, it is doubtful whether LLMs have similar levels of NLG evaluation capability for texts in other languages, especially low-resource languages. As \citep{zhang2024comprehensive} points out, we should be more cautious about using LLMs to evaluate texts in non-Latin languages. We believe that the lack of evaluation capability of LLM-based evaluators on low-resource languages may be due to the insufficient presence of these languages in the pretraining corpus. Therefore, further fine-tuning on certain low-resource languages may be a potential strategy to address this issue, and \citet{DBLP:conf/naacl/HadaGABS24} have already shown promising preliminary results. Additionally, existing research mainly focuses on more traditional NLG tasks such as translation, summarization, and dialogue. However, there are many new scenarios in reality with different requirements and evaluation criteria. For example, using LLMs to automatically evaluate scientific reviews could be valuable in identifying and flagging content that is unfaithful or unclear, alerting reviewers to potential issues. Research on low-resource languages and new task scenarios will provide a more comprehensive understanding of LLMs' evaluation capabilities.

\textbf{Diverse forms of human-LLM collaborative NLG evaluation.} According to the literature reviewed above, there is little research on collaborative evaluation between humans and LLMs. Neither humans nor LLMs are perfect, and each has its strengths. Since the ultimate goal of NLG research is to evaluate text quality more accurately and efficiently, we believe that collaboration between humans and LLMs can achieve better results than pure human evaluation or automatic evaluation. In the collaboration between humans and LLMs, technologies in the field of human-computer interaction may bring new implementation methods to the collaboration. In addition, what roles humans and LLMs should play in the evaluation and how they can better complement each other are still worth researching.

\vfill
\pagebreak

\begin{acknowledgments}
This work was supported by Beijing Science and Technology Program (Z231100007423011) and Key Laboratory of Science, Technology and Standard in Press Industry (Key Laboratory of Intelligent Press Media Technology). We appreciate the anonymous reviewers for their helpful comments. Xiaojun Wan is the corresponding author.
\end{acknowledgments}

\color{black}
\starttwocolumn
\bibliography{custom}

\appendix

\end{document}